# ViBERTgrid: A Jointly Trained Multi-Modal 2D Document Representation for Key Information Extraction from Documents

Weihong Lin[1, *], Qifang Gao[2, †, *], Lei Sun[1], Zhuoyao Zhong[1], Kai Hu[3, †],

Qin Ren[2, †], and Qiang Huo[1]

[1] Microsoft Research Asia, Beijing, China
[2] School of Software and Microelectronics, Peking University, Beijing, China
[3] Department of EEIS., University of Science and Technology of China, Hefei, China
{weihlin, v-qgao, lsun, zhuoyao.zhong, v-kahu1, v-qinren, qianghuo}@microsoft.com

**Abstract.** Recent grid-based document representations like BERTgrid allow the simultaneous encoding of the textual and layout information of a document in a 2D feature map so that state-of-the-art image segmentation and/or object detection models can be straightforwardly leveraged to extract key information from documents. However, such methods have not achieved comparable performance to state-of-the-art sequence- and graph-based methods such as LayoutLM and PICK yet. In this paper, we propose a new multi-modal backbone network by concatenating a BERTgrid to an intermediate layer of a CNN model, where the input of CNN is a document image and the BERTgrid is a grid of word embeddings, to generate a more powerful grid-based document representation, named ViBERTgrid. Unlike BERTgrid, the parameters of BERT and CNN in our multi-modal backbone network are trained jointly. Our experimental results demonstrate that this joint training strategy improves significantly the representation ability of ViBERTgrid. Consequently, our ViBERTgrid-based key information extraction approach has achieved state-of-the-art performance on real-world datasets.

## 1 Introduction

Key Information Extraction (KIE) is the task of extracting the values of a number of pre-defined key fields from documents such as invoices, purchase orders, receipts, contracts, financial reports, insurance quotes, and many more, which is illustrated in Figure 1. KIE is an essential technology for many large-scale document processing scenarios such as fast indexing, efficient archiving, automatic financial auditing and so on.

---

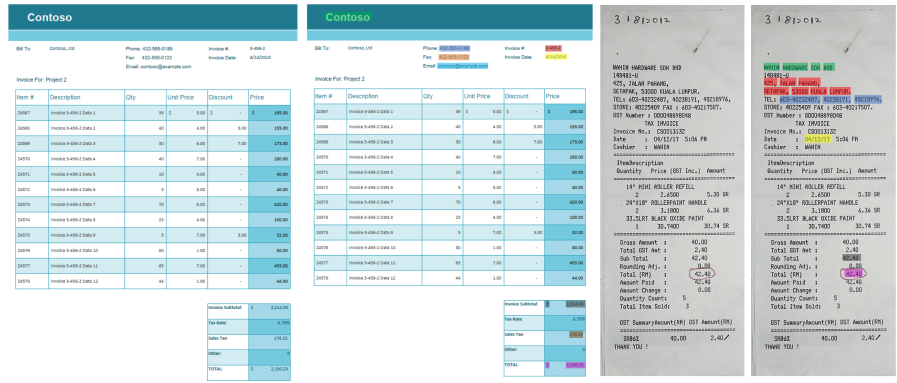

**Figure 1**. Two examples of KIE task. Some key fields, including company name, address, phone number, date, email, fax number, invoice number, subtotal, tax and total amount, are extracted from a fake invoice[1] and a receipt from the SROIE [1] dataset.

With the rapid development of deep learning, many deep learning based KIE approaches have emerged and significantly outperformed traditional rule-based and template-based methods in terms of both accuracy and capability. Most of these approaches simply treat KIE as a token classification problem and use different deep learning models to predict the field type of each document token which could be a character, a subword, or a word. According to their used document representation methods, these approaches could be roughly categorized into three types: sequence-based, graph-based, and grid-based. Sequence-based approaches need to serialize a document into a 1D text sequence first, then use existing sequence tagging models in NLP (e.g., [2, 3, 4, 5, 6]) to extract field values. To reduce the influence of serialization, earlier methods [7, 8, 9] tried to encode the 2D position information of tokens into their token embeddings, but they still relied on the accurate serialization of the text segments so that it was hard to apply them to documents with complex layouts where the serialization step is nontrivial. Two recent works, LayoutLM [10] and LAMBERT [11], proposed to add a 2D position embedding to the input token embedding of the BERT model. After being pre-trained on a large-scale document dataset, its output token embedding could capture the spatial relationship among tokens within a document, which makes LayoutLM and LAMBERT less sensitive to the serialization step. Pramanik et al. [12] and LayoutLMv2 [13] extended the idea of LayoutLM further by integrating the image information in the pre-training stage to learn a stronger multimodal document representation. However, these pre-training based methods rely on large datasets and computation resources to achieve good performance. Graph-based approaches model each document page as a graph, in which text segments (words or text-lines) are represented as nodes. The initial representation of each node could combine visual, textual and positional features of its corresponding text segment [14]. Then graph neural networks or self-attention operations [15] are used to propagate information between neighboring nodes in the graph to get a richer representation for each node. After that, some methods like [16] fed these enhanced node embeddings to a classifier to do field type classification directly, while some other methods [14, 17, 18, 19, 20, 21, 22] concatenated each node

---

[1] https://azure.microsoft.com/en-us/services/cognitive-services/form-recognizer/ - features

embedding to all token embeddings in its corresponding text segment, which were then fed into sequence tagging models to extract field values. Recent graph-based methods like PICK [14], TRIE [21] and VIES [22] have achieved superior performance on the public SROIE dataset [1]. Grid-based approaches like Chargrid [23], Dang et al. [24], CUTIE [25], BERTgrid [26] represented each document as a 2D grid of token embeddings and then use standard instance segmentation models to extract field values from the 2D grid. These grid representations have preserved the textual and layout information of documents but missed image texture information. So VisualWordGrid [27] combined these grid representations with 2D feature maps of document images to generate more powerful multimodal 2D document representations, which could simultaneously preserve visual, textual and layout information of documents. However, we find that these grid-based approaches cannot achieve comparable performance against the state-of-the-art methods like LayoutLM, PICK, TRIE and VIES due to two reasons: 1) Methods including Chargrid, Dang et al. [24], CUTIE and VisualWordGrid don't leverage state-of-the-art contextualized word embedding techniques like BERT to extract strong enough token embeddings; 2) Although BERTgrid incorporates BERT into the grid representation, the parameters of pretrained BERT are fixed during model training so that the potential of BERT-based token embedding is not fully exploited.

In this paper, we propose two effective techniques to significantly improve the accuracy of grid-based KIE methods. First, we propose a new multi-modal backbone network by combining the best of BERTgrid and CNN to generate a more powerful grid-based document representation, named ViBERTgrid, to simultaneously encode the textual, layout and visual information of a document in a 2D feature map. Second, we propose a new joint training strategy to finetune the parameters of both BERT and CNN so that the representation ability of ViBERTgrid is significantly improved. Consequently, our ViBERTgrid-based KIE approach has achieved superior performance on real-world datasets.

## 2    Related Works

Key information extraction from documents has been studied for decades. Before the advent of deep learning based methods, early works [28, 29, 30, 31, 32, 33] mainly depended on some rules or human-designed features in known templates, therefore they usually failed on unseen templates and were not scalable in practical applications. With the development of deep learning, significant progress has been made in KIE. As mentioned above, most of modern deep learning based approaches formulate KIE as a token classification problem. In addition to the above-mentioned works, another work [34] proposed a CNN-based method to jointly perform handwritten text detection, transcription and named entity recognition from input document images. Other than this formulation, KIE can also be formulated as other problems. Majumder et al. [35] proposed a representation learning approach to extracting the values of key fields with prior knowledge. For each field, some candidate words were firstly selected. Then, the feature of each word was embedded with its contextual information and the cosine similarity between this embedding and the embedding of target field was calculated as a similarity score. SPADE [36] formulated KIE as a spatial dependency parsing problem. It constructed a dependency graph with text segments and fields as the graph nodes first, then used a decoder to extract field values from identified connectivity between graph nodes. BROS [37] improved SPADE further by proposing a new position encoding method and an area-masking based pre-training objective. Another category of methods [38, 39, 40] adapted sequence-to-sequence models used in other NLP or image captioning tasks to directly predict all the values of key fields without requiring word-level supervision. Our proposed multimodal 2D document representation can be easily integrated into most of these approaches. Exploring the effectiveness of our document representation for other KIE frameworks will be one of our future works.

Other than KIE, grid-based document representations have also been studied in other document understanding tasks. Xiao et al. [41] constructed a 2D text embedding map with sentence embeddings and combined this textual map and visual features with a fully convolutional network for pixel-level segmentation of image regions such as table, section heading, caption, paragraph and so on. Raphaël et al. [42] proposed a multimodal neural model by concatenating a 2D text embedding to an intermediate layer of a CNN model for a more fine-grained segmentation task on historical newspapers. Unlike our approach, these methods didn't leverage state-of-the-art BERT models to generate text embedding maps and the parameters of their text embedding models and CNN models were not jointly trained.

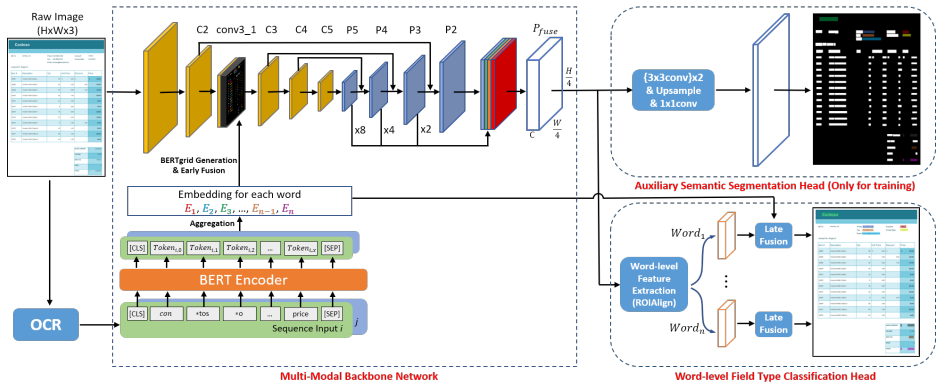

**Figure 2.** Overview of our approach.

## 3 Methodology

As illustrated in Figure 2, our approach is composed of three key components: 1) A multi-modal backbone network used to generate a ViBERTgrid feature map; 2) An auxiliary semantic segmentation head used to perform pixel-wise field type classification; 3) A word-level field type classification head used to predict the field type of each word extracted by an OCR engine. Details of these modules and the proposed joint training strategy will be described in detail in the following subsections.

## 3.1 ViBERTgrid

### 3.1.1 BERTgrid Generation

We follow BERTgrid [26] to construct a 2D word grid, where word embeddings are generated by a pretrained BERT-style language model. Given an input document image, words and their bounding boxes are extracted with an OCR engine. These words are then organized as a unidimensional sequence of length $N$, denoted as $D = (w^{(1)}, w^{(2)}, \cdots, w^{(N)})$, by reading them in a top-left to bottom-right order. The quadrilateral bounding box of $w^{(i)}$ is denoted as $B^{(i)} = [(x_1^i, y_1^i), (x_2^i, y_2^i), (x_3^i, y_3^i), (x_4^i, y_4^i)]$, where $(x_j^i, y_j^i), j \in \{1, 2, 3, 4\}$ represents the coordinates of its corner points in the input document image. Following BERT [5], we tokenize $D$ into a sub-word token sequence of length $M$, denoted as $T = (t^{(1)}, t^{(2)}, \cdots, t^{(M)})$. As the maximum input sequence length of BERT is restricted to 512, we use a sliding window to slice long sequences (with a length greater than 510) into several fixed-length subsequences whose lengths are set to 510. Neighboring subsequences are not overlapped with each other. For each subsequence, two special tokens [CLS] and [SEP] are inserted as the first and the last tokens, respectively. Some [PAD] tokens are added to the end of the last subsequence if its length is less than 512. Then we encode each subsequence with a BERT encoder to get the embedding of each token, denoted as $e(t^{(i)})$, and obtain the word embedding of $w^{(j)}$, denoted as $E(w^{(j)})$, by averaging the embeddings of its tokens. The spatial resolution of our used BERTgrid is $1/S$ of the original document image, where $S$ denotes the stride of the convolutional feature map which is concatenated to. The BERTgrid representation of the document is defined as

$$G_{x,y,:} = \begin{cases} E(w^{(i)}), & \text{if } (x*S, y*S) \prec B^{(i)} \\ 0_d & \text{others} \end{cases} \quad (1)$$

where $\prec$ means "be located in", $d$ is the embedding dimensionality, $0_d$ denotes an all-zero vector used for background whose dimension is $d$, $G$ is a rank-3 tensor, i.e., $G \in R^{(H/S) \times (W/S) \times d}$, H and W are the height and width of the original image, respectively.

### 3.1.2 Visual and Textual Feature Fusion

As shown in Figure 2, we concatenate the generated BERTgrid $G$ to an intermediate layer of a CNN to construct a new multi-modal backbone network. To save computation, a lightweight ResNet18-FPN network [43], which is built on top of a ResNet18-D network [44], is used as the CNN module. ResNet18-D is an improved version of ResNet18 [45]. The output of ResNet18-FPN is a set of down-sampled feature maps, i.e., $P_2, P_3, P_4, P_5$, whose resolutions are 1/4, 1/8, 1/16, 1/32 of the original document image. All of these maps have 256 channels. We set the stride of BERTgrid to 8 by default and conduct an early fusion operation, which first concatenates $G$ to the output of the residual block conv3_1 in ResNet [45] and then performs dimensionality reduction with a 1x1 convolutional layer to make channel number the same as that of CNN feature map. Finally, after resizing $P_3$, $P_4$ and $P_5$ to the size of $P_2$, these four feature maps are concatenated and fed to a 1x1 convolutional layer to generate a feature map $P_{fuse}$ with 256 channels. Here, $P_{fuse}$ is a powerful multi-modal 2D feature map for document understanding tasks.

### 3.2 Word-level Field Type Classification Head

For each word $w^{(i)}$, ROIAlign [46] is used to convert the features inside the word bounding box on the $P_{fuse}$ map into a small $7 \times 7 \times 256$ feature map $X_i$. Then we use an additional skip connection to pass the embedding of each word $E(w^{(i)})$ to the word-level field type classification head so that the error gradients could be propagated backward directly from the output layer to the BERT encoder layers. We find that this skip connection could slightly improve the accuracy of our KIE approach. We use a late fusion module to fuse $X_i$ and $E(w^{(i)})$ for each word. Specifically, each feature map $X_i$ is convolved with two 3x3 convolutional layers to generate a new $7 \times 7 \times 256$ feature map, which is fed into a 1024-d fully connected layer to generate a 1024-d feature vector $x_i$. After that, $x_i$ and $E(w^{(i)})$ are concatenated and fed into an additional fully connected layer to generate a 1024-d new feature vector $x_i'$. As some words could be labeled with more than one field type tags (similar to the nested named entity recognition task), we design two classifiers to input $x_i'$ and perform field type classification for each word. The first classifier is a binary classifier used to predict whether the input word is the value of any pre-defined field. If a word is not pruned by the first classifier, it will be classified again by the second classifier, which is composed of $C$ independent binary classifiers ($C$ is the number of pre-defined field types). The $i^{th}$ binary classifier predicts whether the input word is the value of the $i^{th}$ field, where $i \in [1,2,...,C]$. Each binary classifier is implemented with a Sigmoid function. Let $o_1(w_i)$, $o_{2,k}(w_i)$ denote the output of the first classifier and the $k^{th}$ binary classifier of the second classifier, respectively. The loss $L_1$ of the first classifier is a binary cross-entropy loss computed by

$$L_1 = -\frac{1}{N_1}\sum_i y_1(w_i)\log(o_1(w_i)) + (1-y_1(w_i))\log(1-o_1(w_i)) \quad (2)$$

where $y_1(w_i)$ means the ground truth of $w^{(i)}$ for the first classifier and $N_1$ is the number of words sampled from all the words in this document image. Similar to $L_1$, the loss $L_2$ of the second classifier is as follows:

$$L_2 = -\frac{1}{N_2}\sum_{i,k} y_{2,k}(w_i)\log(o_{2,k}(w_i)) + (1-y_{2,k}(w_i))\log(1-o_{2,k}(w_i)) \quad (3)$$

where $y_{2,k}(w_i)$ is 1 if $w^{(i)}$ has a label of the $k^{th}$ field, and is 0 if this word does not belong to this field, $N_2$ denotes the mini-batch size of the second classifier. We then calculate the total loss $L_C$ of this head by

$$L_C = L_1 + L_2 \quad . \quad (4)$$

### 3.3 Auxiliary Semantic Segmentation Head

While training this end-to-end network, we find that adding an additional pixel-level segmentation loss can make the network converge faster and more stably. Similar to the word-level field type classification head, this semantic segmentation head also includes two classifiers designed for two different classification tasks. The first one is used to classify each pixel into one of three categories: (1) Not located in any word box; (2) Inside a word box which belongs to any pre-defined field; (3) Inside other word boxes. If a pixel is classified into the second category by the first classifier, it will be classified again by the second classifier, which contains $C$ independent binary classifiers for $C$ field types, to get its field type. Specifically, followed by two 3x3 convolutional layers (256 channels), an up-sampling operation and two parallel 1x1 convolutional layers on $P_{fuse}$, two output feature maps $X_1^{out} \in R^{H \times W \times 3}$ and $X_2^{out} \in R^{H \times W \times C}$ are generated. Each pixel on $X_1^{out}$ and $X_2^{out}$ is classified by a Softmax function and $C$ Sigmoid functions to get the results of the first and second classifiers, respectively. Let $o_1'(x, y)$ and $o_{2,k}'(x, y)$ denote the output of the first classifier and the $k^{th}$ binary classifier of the second classifier for a pixel $(x, y)$, respectively. The softmax loss $L_{AUX-1}$ of the first classifier is as follows:

$$L_{AUX-1} = -\frac{1}{N_1'} \sum_{(x,y) \in S_1} CrossEntropyLoss(o_1'(x, y), y_1'(x, y)) \quad (5)$$

where $y_1'(x, y)$ is the ground truth of pixel $(x, y)$ in the first task, $N_1'$ is the number of sampled pixels, $S_1$ is the set of sampled pixels. We calculate the loss $L_{AUX-2}$ of the second task like Equation (3):

$$L_{AUX-2} = -\frac{1}{N_2'} \sum_{(x,y) \in S_2, k} y_{2,k}'(x, y) \log(o_{2,k}'(x, y)) + (1 - y_{2,k}'(x, y)) \log(1 - o_{2,k}'(x, y)) \quad (6)$$

where $y_{2,k}'(x, y) \in \{0, 1\}$ is the ground truth of pixel $(x, y)$ for the $k^{th}$ binary classifier in the second task, $y_{2,k}'(x, y)$ is 1 only if the word, which covers pixel $(x, y)$, has a label of field $k$, $N_2'$ is the mini-batch size of this task, $S_2$ is the set of sampled pixels.

The auxiliary segmentation loss $L_{AUX}$ is the sum of $L_{AUX-1}$ and $L_{AUX-2}$:

$$L_{AUX} = L_{AUX-1} + L_{AUX-2} \quad . \quad (7)$$

### 3.4 Joint Training Strategy

Joint training of a BERT model and a CNN model is not easy as these two types of models are fine-tuned with different optimizers and hyperparameters. Pre-trained BERT models are usually fine-tuned with an Adam optimizer [47] with a very small learning rate, while CNN models are fine-tuned with a standard SGD optimizer with a relatively larger learning rate. So, we train the pre-trained BERT model and the CNN model with an AdamW [48] optimizer and an SGD optimizer, respectively. Hyperparameter settings for these two optimizers are also independent.

Finally, the objective function of our whole network is

$$Loss = L_C + \lambda L_{AUX} \quad (8)$$

where $\lambda$ is a control parameter, which is set to 1 in our experiments.

## 4 Experiments

### 4.1 Dataset

**SROIE**: The ICDAR SROIE dataset is a public dataset for information extraction which contains 626 receipts for training and 347 receipts for testing. It predefines 4 types of entities, including company, date, address and total. Following [10, 14], we use the ground-truth bounding boxes and transcripts of text segments provided by the organizer in our experiments.

**INVOICE**: This dataset is our in-house dataset which contains 24,175 real-world invoice pages for training and 643 invoice pages for testing. These invoices are generated from several hundred templates of complex layout structures. 14 types of key fields are annotated by human labelers, including CustomerAddress, TotalAmount, DueDate, PONumber, Subtotal, BillingAddress, CustomerName, InvoiceDate, TotalTax, VendorName, InvoiceNumber, CustomerID, VendorAddress and ShippingAddress. Statistics of these 14 key fields are listed in Table 1. We extract the transcripts and their corresponding word-level bounding boxes from these invoice images with Microsoft Azure Read API[2].

Table 1. Statistics of INVOICE dataset. All the numbers here mean word numbers.

| Field Name | Training | Testing | Fiel Name | Training | Testing |
|---|---|---|---|---|---|
| Customer Address | 150042 | 4097 | Customer Name | 117567 | 3115 |
| Billing Address | 63108 | 1680 | Vendor Name | 135774 | 3509 |
| Vendor Address | 208544 | 5454 | Customer ID | 32106 | 866 |
| Shipping Address | 18386 | 535 | PO Number | 2878 | 70 |
| Invoice Date | 36407 | 971 | Subtotal | 6692 | 180 |
| Due Date | 25601 | 1442 | Total Tax | 11767 | 323 |
| Invoice Number | 22550 | 601 | Total Amount | 34832 | 955 |

### 4.2 Implementation Details

We implement our approach based on PyTorch and conduct experiments on a workstation with 8 NVIDIA Tesla V100 GPUs (32 GB memory). In training, the parameters of the transformer encoder module and CNN module are initialized with BERT/RoBERTa and ResNet18-D pretrained on the ImageNet classification task, respectively. The parameters of the newly added layers are initialized by using random weights with a Gaussian distribution of mean 0 and standard deviation 0.01. The batch size we use is 16, so that the BatchNorm layers in the CNN module can be trained with Synchronized BatchNorm. As mentioned in Sec. 3.4, we use an AdamW optimizer and an SGD optimizer to optimize the parameters of the BERT encoder and CNN module, respectively. Following the widely used setting of hyperparameters during finetuning, for the AdamW optimizer, the learning rate, betas, epsilon and weight decay are set as 2e-5, (0.9, 0.999), 1e-8 and 1e-2, while for the SGD optimizer, the learning rate, weight decay and momentum are set as 0.016, 5e-4 and 0.9, respectively. All models are trained for 33 epochs and a warmup strategy is applied in the first epoch. The learning rate of SGD is divided by 10 every 15 epochs. In each training iteration, we sample a mini-batch of 64 positive and 64 negative words, a mini-batch of 32 hard positive and 32 hard negative words, a mini-batch of 256 Category-1, 512 Category-2 and 256 Category-3 pixels, a mini-batch of 512 hard positive and 512 hard negative pixels, to calculate $L_1$, $L_2$, $L_{AUX-1}$, $L_{AUX-2}$, respectively. The hard positive and negative words or pixels are sampled with the OHEM [49] algorithm. As each receipt in the SROIE dataset only contains a small number of words, the numbers of sampled words or pixels mentioned above are reduced to half. Note that for many document pages in INVOICE dataset, the number of input sequences can be very large, which could lead to the out-of-memory issue due to the limited memory of GPU. To solve this problem, for each document page, we just randomly select at most $L$ ($L$=10) sequences to propagate gradients so that the memory of intermediate layers belonging to the other sequences can be directly released after acquiring output token embeddings. We adopt a multi-scale training strategy. The shorter side of each image is randomly rescaled to a number from {320, 416, 512, 608, 704} while the longer side should not exceed 800. In inference, we set the shorter side of each testing image as 512.

---

[2] https://docs.microsoft.com/en-us/azure/cognitive-services/computer-vision/concept-recognizing-text

**Table 2.** Performance comparison on INVOICE dataset.

| Entities | BERT-Base | BERT-Large | RoBERTa-Base | RoBERTa-Large | LayoutLM-Base | LayoutLM-Large | BERTgrid | ViBERTgrid (BERT-Base) | ViBERTgrid (RoBERTa-Base) | ViBERTgrid (LayoutLM-Base) |
|---|---|---|---|---|---|---|---|---|---|---|
| CustomerAddress | 87.52 | 90.16 | 89.34 | 91.87 | 91.60 | 92.88 | 91.23 | 92.44 | 93.01 | 92.86 |
| TotalAmount | 89.97 | 92.07 | 92.56 | 93.24 | 92.69 | 93.52 | 91.20 | 92.69 | 93.74 | 94.57 |
| DueDate | 92.52 | 95.75 | 96.38 | 96.23 | 96.89 | 96.64 | 64.10 | 96.71 | 95.64 | 96.38 |
| PONumber | 60.43 | 65.22 | 59.68 | 69.57 | 78.83 | 87.22 | 88.11 | 83.69 | 83.45 | 85.71 |
| Subtotal | 86.83 | 86.13 | 86.83 | 86.86 | 89.32 | 88.46 | 87.08 | 87.57 | 88.52 | 89.73 |
| BillingAddress | 91.24 | 92.47 | 91.76 | 93.99 | 96.02 | 95.68 | 95.52 | 95.87 | 96.91 | 95.70 |
| CustomerName | 88.35 | 89.24 | 88.59 | 88.79 | 89.61 | 89.82 | 89.62 | 90.38 | 91.50 | 91.42 |
| InvoiceDate | 93.21 | 94.70 | 92.40 | 94.33 | 93.04 | 94.23 | 92.75 | 93.70 | 94.27 | 95.49 |
| TotalTax | 90.40 | 91.61 | 90.46 | 91.16 | 92.42 | 90.88 | 85.80 | 92.33 | 92.33 | 93.19 |
| VendorName | 90.76 | 91.99 | 91.06 | 91.91 | 92.30 | 92.49 | 88.62 | 92.58 | 92.99 | 92.45 |
| InvoiceNumber | 88.64 | 90.11 | 88.63 | 87.22 | 90.79 | 91.08 | 89.08 | 91.64 | 91.95 | 91.40 |
| CustomerID | 89.16 | 89.12 | 88.51 | 89.53 | 91.51 | 91.89 | 88.84 | 92.03 | 91.67 | 91.37 |
| VendorAddress | 95.67 | 96.80 | 96.74 | 97.17 | 97.62 | 97.61 | 97.37 | 97.58 | 97.67 | 98.03 |
| ShippingAddress | 80.67 | 89.17 | 82.18 | 87.05 | 92.66 | 91.64 | 93.61 | 90.87 | 92.71 | 91.43 |
| Micro F1 | 90.77 | 92.46 | 91.79 | 92.82 | 93.51 | 93.81 | 90.93 | 93.79 | 94.24 | 94.22 |
| Macro F1 | 87.53 | 89.61 | 88.22 | 89.92 | 91.81 | 92.43 | 88.78 | 92.15 | 92.60 | 92.84 |
| # of Parameters | 110M | 340M | 125M | 355M | 113M | 343M | 142M | 142M | 157M | 145M |

### 4.3 Comparisons with Prior Arts

In this section, we compare our proposed approach with several state-of-the-art methods on INVOICE and SROIE. On INVOICE, we adopt a word-level F1 score as evaluation metric, while on SROIE, we use the official online evaluation tool to calculate the field-level F1 score directly to make our result comparable to others.

**INVOICE.** We first compare our approach with several sequence tagging methods based on BERT, RoBERTa and LayoutLM. These methods directly use the output word embeddings of the transformer encoder to perform token-level classification. For fairer comparisons, we use the same sliding window strategy to slice the long input sequences. As shown in Table 2, BERT and RoBERTa cannot achieve satisfactory results because they only use textual information, while LayoutLM can outperform them by a big margin owing to the encoding of 2D position embeddings and large-scale pre-training. As a grid-based approach, our proposed ViBERTgrid can encode the textual, layout and visual information of a document in a 2D feature map simultaneously. Even without large-scale pre-training, our approach can achieve a comparable result against LayoutLM-Large when only using BERT-Base as the transformer encoder. When replacing BERT-Base with more powerful RoBERTa-Base or LayoutLM-Base, the performance of our model can be further improved to 92.60% and 92.84% in Macro F1 score, respectively. We also compare our approach with BERTgrid, which is another grid-based approach. In our current implementation, we set the stride of BERTgrid as 1 and use it as the input of ResNet18-FPN network. It can be seen that our approach outperforms BERTgrid significantly by improving the Macro F1 score from 88.78% to 92.15% when using BERT-Base as the transformer encoder. The effectiveness of our proposed ViBERTgrid representation and the joint training strategy is clearly demonstrated.

Table 3. Performance comparison on SROIE.

| Model | Need Domain-specific Pretraining | # of Parameters | F1 Score |
|---|---|---|---|
| LayoutLM-Large[3] [10] | ✓ | 343M | 96.04 |
| PICK [14] | | --- | 96.12 |
| VIES [22] | | --- | 96.12 |
| TRIE [21] | | --- | 96.18 |
| LayoutLMv2-Base [13] | ✓ | 200M | 96.25 |
| LayoutLMv2-Large [13] | ✓ | 426M | 96.61 |
| ViBERTgrid (BERT-Base) | | 142M | 96.25 |
| ViBERTgrid (RoBERTa-Base) | | 157M | 96.40 |

**SROIE.** Since SROIE only provides the ground truth transcriptions of key information, we pre-process the training data to generate labels of each text segment by matching their OCR results with these transcriptions. In inference phase, following [10, 14], we adopt a lexicon built from training data to auto-correct the predicted results. We compare our approach with other most competitive methods, including LayoutLM, LayoutLMv2, PICK, TRIE and VIES. As shown in Table 3, our approach achieves better results than LayoutLM, PICK, TRIE and VIES. Even without using millions of document images to do bi-modal domain-specific pretraining, our ViBERTgrid can still achieve comparable results with LayoutLMv2-Base when using BERT-Base and RoBERTa-Base as text embedding models. Although LayoutLMv2-Large achieves slightly higher accuracy than ViBERTgrid (RoBERTa-Base), it contains much more parameters.

---
[3] This result is from https://github.com/microsoft/unilm/tree/master/layoutlm.

**Table 4.** Effectiveness of joint training strategy. "XXX" means that it does not converge.

| Model | BERT module optimizer | CNN & CLS Head optimizer | Lr-T | Lr-V | Micro F1 | Macro F1 |
|---|---|---|---|---|---|---|
| Fixed T+V | --- | SGD | --- | 0.016 | 91.33 | 89.52 |
| T+V | AdamW | AdamW | 0.016 | 0.016 | XXX | XXX |
| T+V | AdamW | AdamW | 2e-5 | 2e-5 | 93.37 | 91.77 |
| T+V | AdamW | AdamW | 2e-5 | 0.016 | 91.89 | 88.48 |
| T+V | SGD | SGD | 0.016 | 0.016 | 89.97 | 87.72 |
| T+V | SGD | SGD | 2e-5 | 2e-5 | 76.84 | 61.82 |
| T+V | SGD | SGD | 2e-5 | 0.016 | 91.95 | 90.15 |
| T+V | AdamW | SGD | 2e-5 | 0.016 | **93.79** | **92.15** |

### 4.4 Ablation Studies

In this section, we conduct a series of ablation experiments to evaluate the effectiveness of each module of our approach and explore the impacts of different experimental settings. Since experimental results tend to be more stable on the much larger INVOICE dataset, all experiments here are conducted on it.

**Effectiveness of joint training strategy**. We compare our joint training strategy with several other training strategies to demonstrate its effectiveness. The results are shown in Table 4. "Fixed T+V" means the parameters of the BERT encoder are fixed during finetuning, which can be considered as a re-implementation version of VisualWordGrid by using BERT. "T+V" means that the parameters of BERT and CNN are trained jointly. "Lr-T" and "Lr-S" are the learning rates of BERT and CNN, respectively. The experimental results demonstrate that our proposed joint training strategy, namely training the BERT model and the CNN model with an AdamW optimizer (learning rate 2e-5) and an SGD optimizer (learning rate 0.016) respectively, can significantly improve the representation ability of grid-based document representations and achieve the best performance among all these joint training strategies.

**Effect of multi-modality features on the KIE task.** We conduct the following ablation study to further examine the contributions of visual and textual features to the KIE task and present the results in Table 5. To remove visual information in our ViBERTgrid, we replace the input image with a zero tensor whose shape is (H, W, 3). To use visual information only, we remove the BERTgrid module from ViBERTgrid and use the remaining CNN layers to generate a 2D feature map directly. The results show that textual information is more important than visual information for the KIE task and combining both visual and textual information achieves better results than using visual or textual information only no matter whether the joint training strategy is used or not.

Table 5. Effect of multi-modality features on the KIE task.

| Use Visual Information | Use Textual Information | Joint Training | Micro F1 | Macro F1 |
|---|---|---|---|---|
| ✓ | | | 87.51 | 83.78 |
| | ✓ | | 91.06 | 89.19 |
| ✓ | ✓ | | 91.33 | 89.52 |
| | ✓ | ✓ | 93.52 | 91.68 |
| ✓ | ✓ | ✓ | **93.79** | **92.15** |

Table 6. Effectiveness of CNN module. BERT Mini [50] is a pretrained miniature BERT with 4 layers, 4 heads and 256 hidden embedding size.

| Model | Micro F1 | Macro F1 |
|---|---|---|
| BERT-Mini | 86.98 | 81.96 |
| BERT-Mini + CNN | **91.36** | **89.00** |
| BERT-Base | 92.47 | 89.02 |
| BERT-Base + CNN | **93.79** | **92.15** |
| RoBERTa-Base | 92.87 | 89.84 |
| RoBERTa-Base + CNN | **94.24** | **92.60** |
| LayoutLM-Base | 94.03 | 92.49 |
| LayoutLM-Base + CNN | **94.22** | **92.84** |

**Effectiveness of the CNN module**. To evaluate the effectiveness of the CNN module, we exclude it and directly use output word embeddings of the transformer encoder to do word-level classification while the training strategies described in Sec. 3.2 are used. As shown in Table 6, no matter which transformer encoder is used, the CNN module can consistently improve the performance. The improvement becomes more significant when the transformer encoder is relatively weaker, e.g., when BERT-Mini is used as the transformer encoder, the CNN module can improve the Macro F1 score by 7.04% absolutely.

Table 7. Comparisons of early fusion at different feature stages.

| Stride of $G$ | Feature Stage | Micro F1 | Macro F1 |
|---|---|---|---|
| 4 | C2 | 93.34 | 91.27 |
| 8 | C3 | **93.79** | **92.15** |
| 16 | C4 | 93.41 | 91.69 |

**Comparison of early fusion at different feature stages**. We train three models by fusing BERTgrid $G$ with the output of the residual block conv2_1 (C2), conv3_1 (C3) and conv4_1 (C4) in ResNet-18D, whose strides are 4, 8, 16, respectively. As shown in Table 7, the second model which fuses $G$ with C3 achieves the best performance.

**Effectiveness of early fusion and late fusion**. In this section, we train different models by using early fusion, late fusion, and both. Moreover, to evaluate the impact of the representation ability of different text embeddings, we construct BERTgrid $G$ from the $3^{rd}$ layer, $6^{th}$ layer and $12^{th}$ layer of BERT-Base, respectively. As shown in Table 8, the combination of early fusion and late fusion can achieve better performance than early fusion or late fusion only under all experimental settings. Furthermore, compared with late fusion, early fusion plays a more important role when the text embeddings are relatively weaker.

**Table 8.** Effectiveness of early fusion ("Early") and late fusion ("Late"). BERT-Base is used as the transformer encoder here.

| Layer Number | Fusion Mode | Micro F1 | Macro F1 |
|---|---|---|---|
| 12 | Early | 93.64 | 91.97 |
| 12 | Late | 93.54 | 91.71 |
| 12 | Early+Late | **93.79** | **92.15** |
| 6 | Early | 93.00 | 91.40 |
| 6 | Late | 92.95 | 91.21 |
| 6 | Early+Late | **93.40** | **92.02** |
| 3 | Early | 91.88 | 90.26 |
| 3 | Late | 90.52 | 87.55 |
| 3 | Early+Late | **92.17** | **90.67** |

## 5 Conclusion and Future Work

In this paper, we propose ViBERTgrid, a new grid-based document representation for the KIE task. By combining the best of BERTgrid and CNN, and training the parameters of BERT and CNN jointly, the representation ability of our ViBERTgrid is much better than other grid-based document representations such as BERTgrid and VisualWordGrid. Consequently, our ViBERTgrid-based KIE approach has achieved state-of-the-art performance on two real-world datasets we experimented.

The proposed ViBERTgrid can be easily integrated into some other KIE frameworks introduced in Sec. 2. Therefore, exploring the effectiveness of our document representation for other KIE frameworks will be one of our future works. Moreover, we will explore the effectiveness of our ViBERTgrid-based document representation for other document understanding tasks where both visual and textual information are useful, such as layout analysis and table structure recognition.

# References

[1] Z. Huang, K. Chen, J. He, X. Bai, D. Karatzas, S. Lu and C. V. Jawahar, "ICDAR2019 Competition on Scanned Receipt OCR and Information Extraction," in *ICDAR*, 2019, pp. 1516-1520.

[2] G. Lample, M. Ballesteros, S. Subramanian, K. Kawakami and C. Dyer, "Neural Architectures for Named Entity Recognition," in *NAACL*, 2016, pp. 260–270.

[3] J. P. Chiu and E. Nichols, "Named Entity Recognition with Bidirectional LSTM-CNNs," *TACL,* vol. 4, pp. 357-370, 2016.

[4] X. Ma and E. Hovy, "End-to-end Sequence Labeling via Bi-directional LSTM-CNNs-CRF," in *ACL*, 2016, pp. 1064–1074.

[5] J. Devlin, M.-W. Chang, K. Lee and K. Toutanova, "BERT: Pre-training of Deep Bidirectional Transformers for Language Understanding," in *NAACL*, 2019, pp. 4171–4186.

[6] Y. Liu, M. Ott, N. Goyal, J. Du, M. Joshi, D. Chen, O. Levy, M. Lewis, L. Zettlemoyer and V. Stoyanov, "RoBERTa: A Robustly Optimized BERT Pretraining Approach," in *arXiv preprint arXiv:1907.11692*, 2019.

[7] R. B. Palm, O. Winther and F. Laws, "CloudScan - A configuration-free invoice analysis system using recurrent neural networks," in *ICDAR*, 2017, pp. 406-413.

[8] C. Sage, A. Aussem, H. Elghazel, V. Eglin and J. Espinas, "Recurrent Neural Network Approach for Table Field Extraction in Business Documents," in *ICDAR*, 2019, pp. 1308-1313.

[9] W. Hwang, S. Kim, M. Seo, J. Yim, S. Park, S. Park, J. Lee, B. Lee and H. Lee, "Post-OCR parsing: building simple and robust parser via BIO tagging," in *Workshop on Document Intelligence at NeurIPS*, 2019.

[10] Y. Xu, M. Li, L. Cui, S. Huang, F. Wei and M. Zhou, "LayoutLM: Pre-training of Text and Layout for Document Image Understanding," in *SIGKDD*, 2020, pp. 1192–1200.

[11] Ł. Garncarek, R. Powalski, T. Stanisławek, B. Topolski, P. Halama, M. Turski and F. Graliński, "LAMBERT: Layout-Aware (Language) Modeling for information extraction," in *arXiv preprint arXiv:2002.08087*, 2020.

[12] S. Pramanik, S. Mujumdar and H. Patel, "Towards a Multi-modal, Multi-task Learning based Pre-training Framework for Document Representation Learning," in *arXiv preprint arXiv:2009.14457*, 2020.

[13] Y. Xu, Y. Xu, T. Lv, L. Cui, F. Wei, G. Wang, Y. Lu, D. Florencio, C. Zhang, W. Che, M. Zhang and L. Zhou, "LayoutLMv2: Multi-modal Pre-training for Visually-Rich Document Understanding," in *arXiv preprint arXiv:2012.14740*, 2020.

[14] W. Yu, N. Lu, X. Qi, P. Gong and R. Xiao, "PICK: Processing Key Information Extraction from Documents using Improved Graph Learning-Convolutional Networks," in *ICPR*, 2020.

[15] A. Vaswani, N. Shazeer, N. Parmar, J. Uszkoreit, L. Jones, A. N. Gomez, L. Kaiser and I. Polosukhin, "Attention Is All You Need," in *NeurIPS*, 2017, pp. 6000–6010.

[16] D. Lohani, A. Belaïd, and Y. Belaïd, "An Invoice Reading System Using a Graph Convolutional Networks," in *ACCV 2018 Workshops*, 2018.

[17] Y. Qian, E. Santus, Z. Jin, J. Guo and R. Barzilay, "GraphIE: A Graph-Based Framework for Information Extraction," in *NAACL*, 2019, pp. 751–761.

[18] X. Liu, F. Gao, Q. Zhang and H. Zhao, "Graph Convolution for Multimodal Information Extraction from Visually Rich Documents," in *NAACL*, 2019, pp. 32–39.

[19] M. Wei, Y. He and Q. Zhang, "Robust Layout-aware IE for Visually Rich Documents with Pre-trained Language Models," in *ACM SIGIR*, 2020, pp. 2367-2376.

[20] C. Luo, Y. Wang, Q. Zheng, L. Li, F. Gao and S. Zhang, "Merge and Recognize: A Geometry and 2D Context Aware Graph Model for Named Entity Recognition from Visual Documents," in *TextGraphs Workshop at COLING*, 2020, pp. 24-34.

[21] P. Zhang, Y. Xu, Z. Cheng, S. Pu, J. Lu, L. Qiao, Y. Niu and F. Wu, "TRIE: End-to-End Text Reading and Information Extraction for Document Understanding," in *ACM Multimedia*, 2020, pp. 1413–1422.

[22] J. Wang, C. Liu, L. Jin, G. Tang, J. Zhang, S. Zhang, Q. Wang, Y. Wu and M. Cai, "Towards Robust Visual Information Extraction in Real World: New Dataset and Novel Solution," in *AAAI*, 2021.

[23] A. R. Katti, C. Reisswig, C. Guder, S. Brarda, S. Bickel, J. Höhne and J. B. Faddoul, "Chargrid: Towards Understanding 2D Documents," in *EMNLP*, 2018, pp. 4459-4469.

[24] T. N. Dang and D. N. Thanh, "End-to-End Information Extraction by Character-Level Embedding and Multi-Stage Attentional U-Net," in *BMVC*, 2019.

[25] X. Zhao, E. Niu, Z. Wu and X. Wang, "CUTIE: Learning to Understand Documents with Convolutional Universal Text Information Extractor," in *arXiv preprint arXiv:1903.12363*, 2019.

[26] T. I. Denk and C. Reisswig, "BERTgrid: Contextualized Embedding for 2D Document Representation and Understanding," in *Document Intelligence Workshop at NeurIPS*, 2019.

[27] M. Kerroumi, O. Sayem and A. Shabou, "VisualWordGrid: Information Extraction From Scanned Documents Using A Multimodal Approach," in *arXiv preprint arXiv:2010.02358*, 2020.

[28] A. Dengel and B. Klein, "smartFIX: A Requirements-Driven System for Document Analysis and Understanding," in *DAS*, 2002, pp, 433-444.

[29] F. Cesarini, E. Francesconi, M. Gori and G. Soda, "Analysis and Understanding of Multi-Class Invoices," *IJDAR,* vol. 6, pp. 102-114, 2003.

[30] E. Medvet, A. Bartoli and G. Davanzo, "A probabilistic approach to printed document understanding," *IJDAR,* vol. 14, pp. 335-347, 2011.

[31] D. Esser, D. Schuster, K. Muthmann, M. Berger and A. Schill, "Automatic Indexing of Scanned Documents - a Layout-based Approach," in *DRR*, 2012.

[32] D. Schuster, K. Muthmann, D. Esser, A. Schill, M. Berger, C. Weidling, K. Aliyev and A. Hofmeier, "Intellix -- End-User Trained Information Extraction for Document Archiving," in *ICDAR*, 2013, pp. 101-105.

[33] M. Rusinol, T. Benkhelfallah and V. Poulain d'Andecy, "Field Extraction from Administrative Documents by Incremental Structural Templates," in *ICDAR*, 2013, pp. 1100-1104.

[34] M. Carbonell, A. Fornés, M. Villegas and J. Lladós, "A Neural Model for Text Localization, Transcription and Named Entity Recognition in Full Pages," *Pattern Recognition Letters,* vol. 136, pp. 219-227, 2020.

[35] B. P. Majumder, N. Potti, S. Tata, J. B. Wendt, Q. Zhao and M. Najork, "Representation Learning for Information Extraction from Form-like Documents," in *ACL*, 2020, pp. 6495-6504.

[36] W. Hwang, J. Yim, S. Park, S. Yang and M. Seo, "Spatial Dependency Parsing for Semi-Structured Document Information Extraction," in *arXiv preprint arXiv:2005.00642*, 2020.

[37] D. K. M. J. Teakgyu Hong, W. Hwang, D. Nam and S. Park, "BROS: A Pre-trained Language Model for Understanding Texts in Document," In Submitted to ICLR, 2021. URL https://openreview.net/forum?id=punMXQEsPr0.

[38] R. B. Palm, F. Laws and O. Winther, "Attend, Copy, Parse - End-to-end information extraction from documents," in *ICDAR*, 2019, pp. 329-336.

[39] H. Guo, X. Qin, J. Liu, J. Han, J. Liu and E. Ding, "EATEN: Entity-aware Attention for Single Shot Visual Text Extraction," in *ICDAR*, 2019, pp. 254-259.

[40] C. Sage, A. Aussem, V. Eglin, H. Elghazel and J. Espinas, "End-to-End Extraction of Structured Information from Business Documents with Pointer-Generator Networks," in *SPNLP Workshop at EMNLP*, 2020, pp. 43-52.

[41] X. Yang, E. Yumer, P. Asente, M. Kraley, D. Kifer and C. L. Giles, "Learning to Extract Semantic Structure from Documents Using Multimodal Fully Convolutional Neural Network," in *CVPR*, 2017, pp. 4342-4351.

[42] R. Barman, M. Ehrmann, S. Clematide, S. A. Oliveira and F. Kaplan, "Combining Visual and Textual Features for Semantic Segmentation of Historical Newspapers," *Journal of Data Mining & Digital Humanities,* HistoInformatics, jdmdh:7097, 2021.

[43] T.-Y. Lin, P. Dollár, R. Girshick, K. He, B. Hariharan and S. Belongie, "Feature Pyramid Networks for Object Detection," in *CVPR*, 2017, pp. 2117-2125.

[44] T. He, Z. Zhang, H. Zhang, Z. Zhang, J. Xie and M. Li, "Bag of Tricks for Image Classification with Convolutional Neural Networks," in *CVPR*, 2019, pp. 558-567.

[45] K. He, X. Zhang, S. Ren and Jian, "Deep residual learning for image recognition," in *CVPR*, 2016, pp. 770-778.

[46] K. He, G. Gkioxari, P. Dollár and R. Girshick, "Mask R-CNN," in *ICCV*, 2017, pp. 2961-2969.

[47] D. P. Kingma and J. Ba, "Adam: A Method for Stochastic Optimization," in *ICLR*, 2015.

[48] I. Loshchilov and F. Hutter, "Decoupled Weight Decay Regularization," in *ICLR*, 2019.

[49] A. Shrivastava, A. Gupta and R. Girshick, "Training Region-based Object Detectors with Online Hard Example Mining," in *CVPR*, 2016, pp. 761-769.

[50] I. Turc, M.-W. Chang, K. Lee and K. Toutanova, "Well-Read Students Learn Better: On the Importance of Pre-training Compact Models," in *arXiv preprint arXiv:1908.08962*, 2019.